# ТРЕХЭТАПНАЯ КОЛИЧЕСТВЕННАЯ НЕЙРОСЕТЕВАЯ МОДЕЛЬ ЯВЛЕНИЯ «НА КОНЧИКЕ ЯЗЫКА»

## П.М.Гопыч

*Харьковский национальный университет им. В.Н.Каразина,*
*пл. Свободы 4, Харьков 61077 Украина, pmg@kharkov.com*

АННОТАЦИЯ

Описана нейросетевая модель явления «на кончике языка» и показано, что оно имеет статистическую природу. Предложены способ вычисления силы и вероятности появления таких состояний, механизм возникновения «чувства знания». Модель объединяет психолингвистический подход и подходы, основанные на анализе памяти и метапамяти, связывает исследовательские традиции, основанные на анализе речевых ошибок и хронометрии наименований. Проанализирован случай на кончике языка из рассказа А.П.Чехова «Лошадиная фамилия». Определен новый «эффект всплескивания руками».

ABSTRACT

A new three-stage computer artificial neural network model of the tip-of-the-tongue phenomenon is shortly described, and it stochastic nature was demonstrated. A way to calculate strength and appearance probability of tip-of-the-tongue states, neural network mechanism of feeling-of-knowing phenomenon are proposed. The model synthesizes memory, psycholinguistic, and metamemory approaches, bridges speech errors and naming chronometry research traditions. A model analysis of a tip-of-the-tongue case from Anton Chekhov's short story 'A Horsey Name' is performed. A new 'throw-up-one's-arms effect' is defined.

ВЕДЕНИЕ

Явление «на кончике языка» (НКЯ) в научной литературе впервые описал У.Джеймс [1]. Оно проявляется как одно из свойств памяти: в состоянии НКЯ человек чувствует, что знает вспоминаемое слово и уверен, что может его вспомнить, хотя в данный момент оно ему временно недоступно. Большинство людей переживают НКЯ (обычно при вспоминании редко встречающихся или мало знакомых слов) самое меньшее один раз в неделю [2], поэтому оно хорошо знакомо практически каждому. Метод количественного исследования явления НКЯ в лабораторных (т.е. контролируемых) условиях предложен в 1966 году [3], и с этого времени оно интенсивно изучается и экспериментально, и теоретически.

В настоящее время при анализе речевого воспроизведения слов сосуществуют две исследовательские традиции [4]: основанная на исследовании речевых ошибок и на хронометрии наименования изображений. Модели явления НКЯ возникли из традиции по исследованию речевых ошибок и развиваются в рамках трех исследовательских подходов [5]: психолингвистического, метакогнитивного и основанного на исследованиях памяти. В [6] на основе модели памяти [7] предложена трехэтапная количественная нейросетевая модель явления НКЯ, объединяющая обе указанные традиции и все три упомянутые выше подхода.

ОСНОВЫ ТРЕХЭТАПНОЙ НЕЙРОСЕТЕВОЙ МОДЕЛИ СОСТОЯНИЙ НКЯ

Существующие компьютерные модели воспроизведения речи являются нейросетевыми моделями, узлы которых представляют собой целостные лингвистические единицы [4]. Согласно нашей модели [6] каждый словарный узел строится из нескольких взаимосвязанных обученных двухслойных автоассоциативных искусственных нейронных сетей (ИНС), каждая из которых хранит отдельные семантические, лексические или фонологические компоненты слова (такое многомерное представление элементов словарной памяти согласуется с данными картирования распределений активности человеческого мозга при обработке им речевой информации [8]). Упомянутые ИНС [6,7] оперируют с наборами (пакетами, образами) двоичных сигналов, кодирующих лингвистическую информацию (это наборы положительных и отрицательных единиц, моделирующих пакеты нервных импульсов, одновременно воздействующих на возбудительные и тормозные синапсы живых нейронов).

Формирование и разрешение состояний НКЯ проходит три основных этапа [6].

I. *Локализация словарного узла,* когда на основе предоставляемой семантической информации выделяется и частично активируется (помечается) группа нейронов, содержащая информацию, относящуюся к вспоминаемому слову (это элементы памяти в виде выше упомянутых ИНС). Здесь же каждой из выделенных ИНС сопоставляется соответствующий ей в метапамяти эталонный вектор [7], который необходим при последующем оценивании адекватности результатов вспоминания (модельный механизм формирования эталонного следа в метапамяти – предмет отдельного обсуждения).

II. *Извлечение слова (компоненты словарного узла) из памяти* (когнитивный процесс предметного уровня), когда путем тестирования (по правилам из [7]) всех взаимосвязанных сетей выделенного узла случайными (свободное вспоминание) или отчасти случайными (вспоминание с подсказкой) входными векторами, извлекается содержащаяся в них информация. Всплывающий на выходе каждой тестируемой сети ее выходной вектор (набор положительных и отрицательных единиц) подлежит сравнению со своим для каждой из этих сетей эталоном, который был локализован на этапе I.

III. *Сравнение* образа, возникшего на выходе ИНС на данном шаге вспоминания, с эталонным образом (вектором) из метапамяти и *принятие решения* о прекращении или продолжении вспоминания (это процессы метакогнитивного уровня). Если для некоторой сети всплывший и эталонный векторы совпадают, то для нее вспоминание завершено. В противном случае этап II повторяется, другой случайный (или отчасти случайный) вектор поступает на вход той же выделенной ИНС и т.д. пока эталонный образ не будет идентифицирован или серия попыток вспоминания не будет остановлена по независимым внешним причинам (например, из-за исчерпания заданного лимита повторений). Такая серия попыток (шагов) вспоминания до исчерпания лимита повторений выполняется преимущественно без вмешательства со стороны сознания.

Собственно извлечение информации из памяти происходит на этапе II, поэтому количественные характеристики этого процесса определяются (когда сеть и эталонный вектор заданы) только свойствами тестируемой сети. Рассмотрим их подробнее на примере (рис.1).

Согласно [7] кривые 0 на рис.1a и рис.1b описывают характеристики вспоминания для случаев, когда обучение идеально и отсутствует забывание. Кривые 1-3 описывают вероятности вспоминания для случаев, когда обучение идеально, но имеет место частичное повреждение (забывание) хранимой сетью информации за счет потери межнейронных связей (рис.1a) или входных нейронов (рис.1b). Кривые 0,1,2,3 на рис.1a и 0,1,2 на рис.1b однотипны: их производные равны нулю при $d=0$ (узнавание) и близких значениях $d$. В отличие от этого для кривой 3 на рис.1b при $d=0$ производная велика (по модулю), что соответствует быстрому изменению вероятности вспоминания при больших подсказках *1-d* (область *узнавания* запомненной информации), а близость нулю этой производной имеет место только при бо́льших *d*. Именно кривая 3 на рис.1b дает пример зависимости *P(d)*, соответствующей, согласно модели [6], состоянию НКЯ.

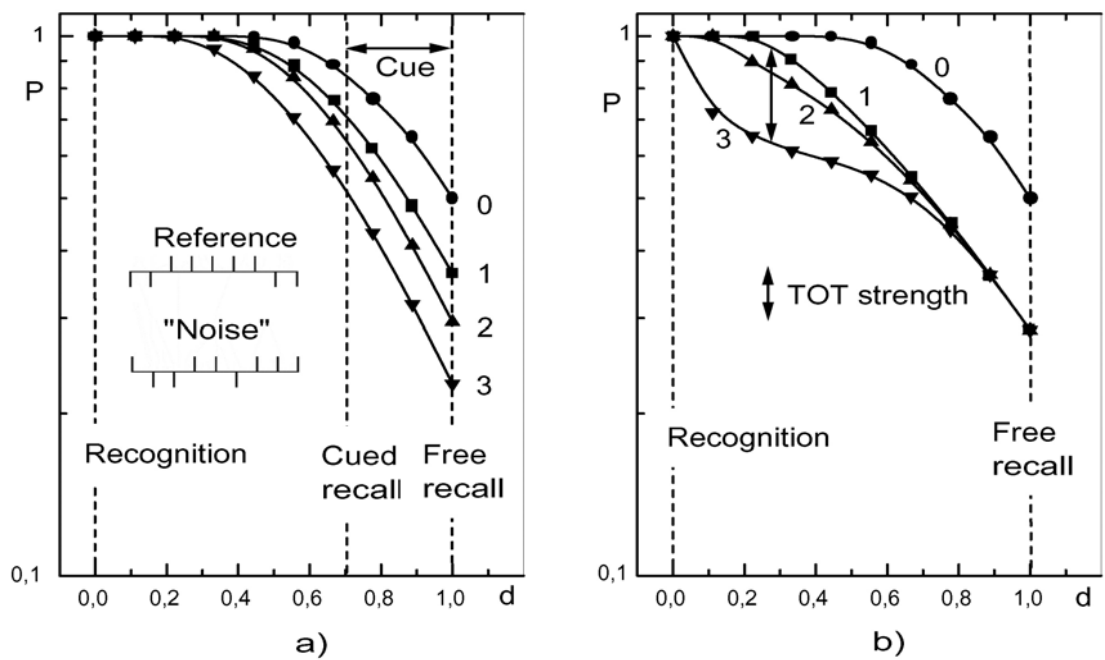

Рис.1а. Зависимость вероятности *P(d)* вспоминания запомненного сетью эталона от степени его искажения *d*. Кривые соединяют значения *P(d)*, вычисленные по правилам из [7] (разные значки) для сети без искажений (кривая 0) и для той же сети, но с 10 случайно выбранными и разорванными межнейронными связями (кривые 1,2,3). Расчет для ИНС с межнейронными связями типа «все со всеми», синаптической матрицей [7] и числом входных (выходных) нейронов *N=9*. Схематически показаны эталонный (Reference) и пример шумового ("Noise") векторов (черточки вверх - проекции '+1', черточки вниз – проекции '–1'). Штриховыми линиями отмечены значения *d*, соответствующие узнаванию эталона (*d =0*), примеру вспоминания с подсказкой *q=1-d=0,3* (*d=0,7*) и свободному вспоминанию (*d =1*).

Рис.1b. То же для сети без повреждений (кривая 0) и для той же сети, но с четырьмя случайно отобранными входными нейронами, которые, полагаем, погибли (или отсутствуют по другой причине) со всеми своими межнейронными связями (кривые 1,2,3). Стрелка схематически обозначает силу состояния НКЯ, описываемого кривой 3. Условия вычислений и другие обозначения см. в подписи к рис.1а.

В качестве оценки силы («глубины») переживаемого состояния НКЯ принимаем степень различия между кривыми 1 и 3 на рис.1b (это может быть, например, максимальное значение разности между ними). Вероятность возникновения состояния НКЯ можно оценить, вычислив вероятность реализации кривой 3 на рис.1b при случайном выборе погибших во входном слое нейронов. Для нашего примера, следуя методике вычислений описанной в [7], найдем следующие значения вероятностей: кривая 1 – 46,8%, кривая 2 – 48,4%, кривая 3 – 4,8%. Следовательно, в рамках модели [6] состояния НКЯ имеют существенно статистическую природу, а для рассматриваемого примера вероятность возникновения НКЯ составляет 4,8%. Состояния НКЯ могут возникать (рис.1b), если сетью частично утрачены входные нейроны. При частично утраченных межнейронных связях (рис.1а) состояния НКЯ не возникают. Следовательно, согласно модели [6], причина возникновения состояний НКЯ – неполная (неадекватная) локализация соответствующей нейронной сети из состава словарного узла (на этапе I).

В нашем примере при любом выборе номеров погибших нейронов все поврежденные сети характеризуется (рис.1b) одинаковой и достаточно большой вероятностью свободного вспоминания $P(1)=28,5\%$. Поэтому для таких сетей число повторений этапов II и III, требуемых для успешного вспоминания, может быть меньше упомянутого выше лимита повторений. В этом случае время разрешения НКЯ относительно невелико и упомянутая выше трехэтапная схема описывает такие состояния полностью.

Однако существуют «затяжные» состояния НКЯ, разрешение которых требует много времени. Для них модель [6] предполагает учет дополнительных механизмов метакогнитивного (см. например, [9]) контроля, слежения и управления как на уровне сознания, так и на подсознательном уровне. Тогда при разрешении НКЯ число и последовательность повторений этапов II-III или даже I-II-III может зависеть от предшествующих результатов вспоминания, подсказок и принимаемой стратегии вспоминания.

Число шагов, используемых при вспоминании временно недоступного слова (число предпринимаемых попыток извлечения из памяти), длительности временных интервалов между ними, и (средняя) длительность пакета нервных импульсов, кодирующих информацию на разных этапах вспоминания, определяют длительность (хронометрию) процесса разрешения состояний НКЯ. Следовательно, обсуждаемая модель пригодна для количественного анализа эффектов вспоминания в их динамике и, тем самым, перебрасывает мостик между исследовательскими традициями, основанными на анализе речевых ошибок и на хронометрии наименования зрительных образов.

АНАЛИЗ СЛУЧАЯ НКЯ ИЗ РАССКАЗА А.П.ЧЕХОВА «ЛОШАДИНАЯ ФАМИЛИЯ»

Проанализируем разрешение затяжных состояний НКЯ на примере случая из рассказа А.П.Чехова «Лошадиная фамилия» [10], в котором Иван Евсеевич (И.Е.), приказчик генерала Булдеева, долго не может вспомнить фамилию акцизного, известного способностью избавлять от зубной боли заговором. Как врач точно и достоверно и как писатель художественно совершенно А.П.Чехов выделяет и описывает существенные детали этого выразительного случая нарушения памяти.

Приведем существенные для модели факты из рассказа (впервые опубликован в 1885 году), относящиеся к переживаемому приказчиком (И.Е.) состоянию, и дадим их модельную [6] интерпретацию:

1. И.Е. нарушениями памяти не страдает. Значит, модель может предполагать, что используемые ИНС не повреждены и могут быть локализованы полностью.

2. И.Е. вспомнил ряд фактов и эпизодов, относящиеся к человеку с искомой фамилией: где, кем работал, когда, куда, почему уехал, с кем живет, чем занимается и т.д., помнит часть вспоминаемого полного имени (имя и отчество - Яков Васильевич). Значит, локализация сетей, относящихся к нужному словарному узлу, выполнена (этап вспоминания I) и для тех из них, которые содержат семантические сведения и сведения из эпизодической памяти, информация успешно извлечена (для них успешно завершены этапы вспоминания II, III). Сама же вспоминаемая фамилия остается пока недоступной. Значит, из сети, содержащей фонологические сведения об искомой фамилии, попытки извлечь информацию безуспешны и процесс ее вспоминания пока прерывается по независимым внешним причинам (например, вследствие исчерпания лимита неосознаваемых повторений этапов II-III такого вспоминания).

3. И.Е. уверен, что знает искомую «лошадиную» фамилию. Значит, у него возникло «чувство знания» [9] этой фамилии. Согласно модели [6] необходимой предпосылкой возникновения чувства знания является факт успешной (на этапе I) локализации в метапамяти эталонного вектора для верификации результатов вспоминания при тестировании сети, содержащей вспоминаемую информацию. Осознается же возникшее чувство знания после (первой) неудачной попытки вспоминания (серии повторений этапов II-III) на основе уже имеющейся информации об успешной локализации вектора-эталона [6]. Наличие чувства знания существенно влияет на выбор дальнейшей стратегии вспоминания.

4. И.Е. мучается вспоминанием день, ночь и следующее утро, стимулируемый напоминаниями и премией в пять рублей. Он сам и другие используют в качестве семантической подсказки то, что фамилия «как бы лошадиная». В качестве подсказки И.Е. использует также (отчасти неосознанно) семантическую и эпизодическую информацию, извлеченную ранее при тестировании сетей выделенного словарного узла (пункт 2). Значит, согласно модели [6], под влиянием возникшего чувства знания (пункт 3) и стимулирующих факторов И.Е. выбирает стратегию, которая состоит в продолжении вспоминания с сохранением его условий (с той же семантической подсказкой). Возникает повторяющийся процесс: вначале преимущественно неосознанно-автоматическое вспоминание (серия повторений этапов II-III), а после неудачи каждой серии - осознанно-волевое (под влиянием чувства знания и стимулирующих факторов) принятие решения о повторении вспоминания в тех же условиях. Затяжной характер этого процесса в сочетании с относительно большой модельной вероятностью свободного вспоминания (*28,5%* на рис.1b) вынуждают предположить, что тестируемая сеть первоначально была выделена неверно и требуемая информация из нее вообще не может быть извлечена (т.е. характеристики первоначально выделенной сети на самом деле не описываются кривыми на рис.1).

5. Вспоминание происходит внезапно и сразу при предоставлении подсказки, практически совпадающей со вспоминаемой фамилией (овес – Овсов). В рамках модели [6] это означает, что под влиянием новой подсказки происходит (без участия сознания И.Е.) изменение стратегии вспоминания: оно повторяется полностью, начиная с этапа I. Вначале с использованием новой семантической информации повторно локализуется требуемая сеть (этап I) и тем самым устраняются ошибки ее локализации, затем для этой сети формируется новый входной вектор, почти совпадающий с эталоном. Поступление нового входного вектора на вход обновленной сети сразу (при однократном выполнении этапа II) приводит к возникновению на ее выходе вектора, совпадающего с эталоном, т.е. происходит *узнавание* эталона. Следовательно, *немедленное разрешение затяжного состояния НКЯ происходит благодаря сочетанию двух факторов: уточнению локализации тестируемой сети и почти полному совпадению подсказки с эталоном*. Успешное разрешения состояния НКЯ подтверждает также справедливость возникшего ранее чувства знания и правильность изначальной (до предоставления решающей подсказки) локализации вектора-эталона. Кроме того, проведенное рассмотрение вынуждает предположить, что процесс вспоминания не содержит механизмов независимого промежуточного контроля качества локализации нейронных сетей, содержащих вспоминаемую информацию, и такой контроль осуществляется только путем верификации конечного результата вспоминания.

6. Вспомнив, И.Е. «тупо поглядел», «как-то дико улыбнулся и, не сказав… ни одного слова, всплеснув руками, побежал к усадьбе…», чтобы сообщить генералу найденную фамилию. Значит внезапное разрешение НКЯ после длительного стимулируемого вспоминания сопровождается острой эмоциональной реакцией, которая проявляется вначале преимущественно бессознательно автоматическим (всплеснул руками…), а потом и преимущественно сознательно планируемым (побежал к усадьбе…) поведением. Такую реакцию назовем «эффектом всплескивания руками». Схема его возникновения и разрешения была только что описана.

Приведенная ниже таблица иллюстрирует в упрощенном виде предполагаемое моделью [6] соотношение между содержанием литературного оригинала [10], психолингвистическим, основанным на модели памяти [7] и метакогнитивным [9] аспектами процесса разрешения состояния НКЯ из рассказа А.П.Чехова «Лошадиная фамилия»

(номера при цитировании оригинала на этапе IIа проставлены при составлении таблицы). В таблице совмещены тесно взаимосвязанные, но разные процессы: вспоминание собственно искомой фамилии (этап IIb – вспоминание с подсказкой исходной семантики, этап IIс – свободное вспоминание, этап IId – вспоминание с подсказкой новой семантики) и вспоминание сопутствующей информации, которая хранится в нейронных сетях того же словарного узла (этап IIа). Этап III процесса вспоминания выделен только для случая успешного вспоминания, хотя очевидно, что после этапа II (тестирование сети) этап III (верификация выходного вектора и принятие решения о прекращении/продолжении вспоминания) необходим при каждой попытке вспоминания.

Таблица 1

Психолингвистический, основанный на модели памяти и метакогнитивный аспекты процесса разрешения состояния НКЯ из рассказа А.П.Чехова «Лошадиная фамилия»

| Этап | Краткое содержание этапа разрешения состояния НКЯ | | | |
| --- | --- | --- | --- | --- |
| | Литературный оригинал [10] (цитирование предложений или их частей) | Модельное описание | | |
| | | Психолингвистический аспект [5] | Аспект, основанный на модели памяти [7] | Метакогнитивный аспект [9] |
| 1 | 2 | 3 | 4 | 5 |
| I | У отставного генерал-майора Булдеева разболелись зубы. На предложение вырвать зуб генерал ответил отказом. Все домашние… предлагали каждый свое средство. Приказчик Булдеева… посоветовал лечиться заговором. | Общее семантическое описание ситуации (задание контекстной информации, определяющей условия вспоминания имени знахаря). | Локализация и частичная активация словарного узла (памяти о знахаре и заговаривании зубов). Генерация векторов подсказок и эталонов. | Возникновение чувства знания вспоминаемой фамилии. |
| II а | 1) В нашем уезде… 2) лет десять назад 3) служил акцизным 4) Яков Васильевич… А после того, как 5) его из акцизных уволили, 6) в Саратове у тещи живет. 7) Тамошних, саратовских на дому у себя пользует, а ежели которые из других городов, то по телеграфу. 8) До водки очень охотник, 9) живет не с женой, а с немкой, 10) ругатель… А фамилию вот и забыл!… Давеча, как сюда шел, помнил… Такая еще простая фамилия… словно как бы лошадиная… Помню, фамилия лошадиная, а какая – из головы вышибло… | Свободное, при наличии только контекстной подсказки, вспоминание информации о человеке, который лечит зубы заговором (пункты 1-10). Однако его фамилия остается не найденной. | Тестирование сетей локализованного узла и успешное извлечение запомненной в них информации (пункты 1-10), вспоминание семантики фамилии акцизного. Тестирование сети, содержащей информацию о вспоминаемой фамилии, не приводит к всплыванию вектора, совпадающего с эталоном. | Неявные, автоматические операции, с относительно низкой степенью их осознания (нижний пласт опыта, подсознательный стиль поведения), которые влияют на поведение прямо и автоматически, без посредничества сознания и без осознанного (волевого) контроля. |

| 1 | 2 | 3 | 4 | 5 |
|---|---|---|---|---|
| II b | И в доме, все наперерыв, стали изобретать фамилии. Перебрали все возрасты, полы и породы лошадей, вспомнили гриву, копыта, сбрую… Приказчика то и дело требовали в дом. Нетерпеливый, замученный генерал пообещал дать пять рублей тому, кто вспомнит настоящую фамилию… | Изобретение разных подсказок общей семантики. Продолжение безуспешных попыток усиленного, стимулируемого вспоминания искомой фамилии с использованием этих подсказок. | Сохранение стратегии вспоминания: многократное повторение идентичных попыток тестирования сети, в которой запомнена искомая фамилия акцизного, с использованием подсказок общей семантики. | Явные операции с относительно высоким уровнем их осознания (верхний пласт опыта; осознанный, явно, контролируемый стиль поведения), которые влияют на поведение при посредничестве сознания и при наличии волевого контроля. |
| | Утром генерал опять послал за доктором… Приехал доктор и вырвал больной зуб. Боль утихла тотчас же… Получив, что следует, за труд, доктор сел в свою бричку и поехал домой. | Появление независимого персонажа, который не участвовал во вспоминании и не знает семантики исходной подсказки | - | - |
| II c | За воротами в поле доктор встретил Ивана Евсеевича… Приказчик стоял на краю дороги и, глядя сосредоточенно себе под ноги, о чем-то думал. Судя по морщинам, бороздившим его лоб, и по выражению глаз, думы его были напряженны, мучительны... Он бормотал. | Продолжение попыток свободного, без подсказок, усиленного вспоминания. | Сохранение стратегии вспоминания: свободное, без подсказок, вспоминание искомой фамилии путем многократного тестирования той же выделенной сети. | Продолжение явных операций с относительно высоким уровнем их осознания и их осознанного (волевого) контроля (верхний пласт опыта). |
| II d | - Иван Евсеич! – обратился к нему доктор. – Не могу ли я, голубчик, купить у вас четвертей пять овса? | Предоставление подсказки новой семантики, лишь отчасти связанной с подсказками, использовавшимися ранее. | Изменение стратегии вспоминания: повторная локализация сети, формирование нового входного вектора, тестирование сети новым входным вектором. | Инициализация и выполнение нового цикла неявных, автоматических операций, с относительно низкой степенью их осознания. |

| 1 | 2 | 3 | 4 | 5 |
|---|---|---|---|---|
| III | Иван Евсеич тупо погля-дел на доктора, как-то дико улыбнулся и, не ска-зав в ответ ни одного слова, всплеснув руками, побежал к усадьбе с такой быстротой, точно за ним гналась бешеная собака. - Надумал! – закричал он радостно, не своим голо-сом, влетая в кабинет к генералу. – Надумал, дай бог здоровья доктору! Овсов! Овсов фамилия акцизного! | Успешное вспоминание искомой фами-лии и прекра-щение процесса вспоминания. Осознание фак-та успешного вспоминания, сопровождаю-щееся вначале мало контроли-руемыми, а за-тем осознан-ными действи-ями и острой эмоциональной реакцией. | Сравнение но-вого, всплывше-го на выходе тестируемой сети, выходного вектора с эта-лоном из мета-памяти, обнару-жение факта их совпадения и принятие реше-ния об успеш-ном завершении процесса вспо-минания. | Проявление прямо-го и автоматиче-ского влияния на поведение неявно-автоматических операций с относи-тельно низкой сте-пенью их осозна-ния, а затем прояв-ление осознанного поведения, контро-лируемого волевым образом при по-средничестве соз-нания. |

ЛИТЕРАТУРА